\pdfoutput=1

\def\year{2021}\relax
\documentclass[letterpaper]{article} 
\usepackage{aaai21}  
\usepackage{times}  
\usepackage{helvet} 
\usepackage{courier}  
\usepackage[hyphens]{url}  
\usepackage{graphicx} 
\usepackage{natbib}  
\usepackage{caption} 
\usepackage{amssymb,amsmath}
\usepackage{algorithm}
\usepackage{algorithmic}
\usepackage{epstopdf}
\usepackage{subfigure}

\def \R {\mathbb{R}}

\DeclareMathOperator*{\cols}{cols}

\DeclareMathOperator*{\tr}{tr}
\DeclareMathOperator*{\sr}{sr}

\DeclareMathOperator*{\nnz}{nnz}
\DeclareMathOperator*{\spr}{sprand}
\DeclareMathOperator*{\svd}{SVD}
\DeclareMathOperator*{\qr}{QR}

\DeclareMathOperator*{\pr}{Pr}

\DeclareMathOperator*{\dcod}{DS}
\DeclareMathOperator*{\bsi}{VSI}
\DeclareMathOperator*{\si}{SI}

\newtheorem{thm}{Theorem}
\newtheorem{lem}{Lemma}

\title{Approximate Multiplication of Sparse Matrices with Limited Space}
\author{
    Yuanyu Wan,~~Lijun Zhang\thanks{Lijun Zhang is the corresponding author.}
}
\affiliations{

    National Key Laboratory for Novel Software Technology,\\
    Nanjing University, Nanjing 210023, China\\
    \{wanyy, zhanglj\}@lamda.nju.edu.cn
}

\begin{document}
\maketitle

\begin{abstract}
Approximate matrix multiplication with limited space has received ever-increasing attention due to the emergence of large-scale applications. Recently, based on a popular matrix sketching algorithm---frequent directions, previous work has introduced co-occuring directions (COD) to reduce the approximation error for this problem. Although it enjoys the space complexity of $O((m_x+m_y)\ell)$ for two input matrices $X\in\mathbb{R}^{m_x\times n}$ and $Y\in\mathbb{R}^{m_y\times n}$ where $\ell$ is the sketch size, its time complexity is $O\left(n(m_x+m_y+\ell)\ell\right)$, which is still very high for large input matrices. In this paper, we propose to reduce the time complexity by exploiting the sparsity of the input matrices. The key idea is to employ an approximate singular value decomposition (SVD) method which can utilize the sparsity, to reduce the number of QR decompositions required by COD. In this way, we develop sparse co-occuring directions, which reduces the time complexity to $\widetilde{O}\left((\nnz(X)+\nnz(Y))\ell+n\ell^2\right)$ in expectation while keeps the same space complexity as $O((m_x+m_y)\ell)$, where $\nnz(X)$ denotes the number of non-zero entries in $X$ and the $\widetilde{O}$ notation hides constant factors as well as polylogarithmic factors. Theoretical analysis reveals that the approximation error of our algorithm is almost the same as that of COD. Furthermore, we empirically verify the efficiency and effectiveness of our algorithm.
\end{abstract}
\section{Introduction}
Matrix multiplication refers to computing the product $XY^T$ of two matrices $X\in\mathbb{R}^{m_x\times n}$ and $Y\in\mathbb{R}^{m_y\times n}$, which is a fundamental task in many machine learning applications such as regression \cite{Naseem10,Cohen16}, online learning \cite{Hazan_2007,Chu11,Zhang2016,Zhang-NIPS-2017}, information retrieval \cite{Eriksson-Bique11} and canonical correlation analysis \cite{Hotelling36,Chen15}. Recently, the scales of data and models in these applications have increased dramatically, which results in very large data matrices. As a result, it requires unacceptable time and space to directly compute $XY^T$ in the main memory. To reduce both time and space complexities, approximate matrix multiplication (AMM) with limited space, which can efficiently compute a good approximation of the matrix product, has been a substitute and received ever-increasing attention \cite{Qiaomin16,Wan-PAKDD-2018,IJCAI18-Wan,ELB}.

Given two large matrices $X\in\mathbb{R}^{m_x\times n}$ and $Y\in\mathbb{R}^{m_y\times n}$, the goal of AMM with limited space is to find two small sketches $B_X\in\mathbb{R}^{m_x\times \ell}$ and $B_Y\in\mathbb{R}^{m_y\times \ell}$ such that $B_XB_Y^T$ approximates $XY^T$ well, where $\ell\ll\min(m_x,m_y,n)$ is the sketch size. Traditionally, randomized techniques such as column selection \cite{Drineas06} and random projection \cite{Sarlos06,Magen2011,Cohen16} have been utilized to develop lightweight algorithms with the $O(n(m_x+m_y)\ell)$ time complexity and $O((m_x+m_y)\ell)$ space complexity for AMM, and yielded theoretical guarantees for the approximation error. Specifically, early studies \cite{Drineas06,Sarlos06} focused on the Frobenius error, and achieved the following bound
\begin{equation}
\label{F_bound1}
\|XY^T-B_XB_Y^T\|_F\leq\epsilon\|X\|_F\|Y\|_F
\end{equation}
with $\ell=\widetilde{O}(1/\epsilon^2)$. Later, two improvements \cite{Magen2011,Cohen16} established the following error bound measured by the spectral norm
\begin{equation}
\label{S_bound1}
\|XY^T-B_XB_Y^T\|\leq\epsilon\|X\|\|Y\|
\end{equation}
with $\ell=\widetilde{O}\left(\left(\sr(X)+\sr(Y)\right)/\epsilon^2\right)$ where $\sr(X)=\frac{\|X\|_F^2}{\|X\|^2}$ is the stable rank of $X$.

However, their sketch size $\ell$ has a quadratic dependence on $1/\epsilon$, which means that a large sketch size is required to ensure a small approximation error. To improve the dependence on $1/\epsilon$, recent studies \cite{Qiaomin16,CoD17} have extended a deterministic matrix sketching technique called frequent directions (FD) \cite{Liberty2013,Ghashami2014,Ghashami2016} to AMM. Specifically, \citet{Qiaomin16} concatenated $X,Y$ vertically as $[X;Y]$, and applied FD to $[X;Y]\in\mathbb{R}^{(m_x+m_y)\times n}$ to generate $B_X$ and $B_Y$ such that
\begin{equation}
\label{S_bound2-1}
\|XY^T-B_XB_Y^T\|\leq\left(\|X\|_F^2+\|Y\|_F^2\right)/\ell
\end{equation}
which also requires the $O(n(m_x+m_y)\ell)$ time complexity and $O((m_x+m_y)\ell)$ space complexity. Furthermore, \citet{CoD17} proposed a new algorithm named as co-occuring directions (COD) with the $O((m_x+m_y)\ell)$ space complexity to generate $B_X$ and $B_Y$ such that
\begin{equation}
\label{S_bound2}
\|XY^T-B_XB_Y^T\|\leq2\|X\|_F\|Y\|_F/\ell.
\end{equation}
Although the error bound of COD is similar to that in (\ref{S_bound2-1}), \citet{CoD17} showed that COD usually has a better empirical performance. Compared with previous randomized methods, COD only requires $\ell=2\sqrt{\sr(X)\sr(Y)}/\epsilon$ to achieve the error bound in (\ref{S_bound1}), which improves the dependence on $1/\epsilon$ to be linear. However, the time complexity of COD is $O\left(n(m_x+m_y+\ell)\ell\right)$, which is still very high for large matrices.

In this paper, we exploit the sparsity of the input matrices to reduce the time complexity of COD. In many real applications, the sparsity is a common property for large matrices. For example, in information retrieval, the word-by-document matrix could contain less than $5\%$ non-zero entries \cite{Dhillon01}. In recommender systems, the user-item rating matrix could contain less than $7\%$ non-zero entries \cite{Zhang17}. The computational bottleneck of COD is to compute the QR decomposition $O(n/\ell)$ times. We note that a similar bottleneck also exists in FD, which needs to compute SVD $O(n/\ell)$ times. To make FD more efficient for sparse matrices, \citet{SFD16} utilized a randomized SVD algorithm named as simultaneous iteration (SI) \cite{SIPower15} to reduce the number of exact SVD. Inspired by this work, we propose to accelerate COD for sparse matrices by utilizing SI. Our main contributions are summarized as follows.
\begin{itemize}
\item We first develop verified simultaneous iteration (VSI) by introducing a verification process, which can efficiently perform a good decomposition for the product of two small sparse matrices with a sufficiently large probability.
\item Then, we develop sparse co-occuring directions (SCOD) by employing VSI to reduce the number of QR decompositions required by COD. In this way, the time complexity is reduced to $\widetilde{O}\left((\nnz(X)+\nnz(Y))\ell+n\ell^2\right)$ in expectation.
\item Moreover, we prove that the space complexity of our algorithm is still $O((m_x+m_y)\ell)$, and it enjoys almost the same error bound as that of COD.
\end{itemize}

\section{Preliminaries}
In this section, we review necessary preliminaries about co-occuring directions and simultaneous iteration.
\subsection{Co-occuring Directions}
Co-occuring directions \cite{CoD17} is an extension of frequent directions \cite{Liberty2013} for AMM. For brevity, the most critical procedures of COD are extracted and summarized in Algorithm \ref{DenseCoD}, which is named as dense shrinkage (DS), where $\sigma_{x}(A)$ is the $x$-th largest singular value of any matrix $A$, and $\max(A,0)$ is performed on each entry of $A$. Given ${X}\in\R^{m_x\times n}$ and ${Y}\in\R^{m_y\times n}$, COD first initializes ${B_X}={0}_{m_x\times \ell}$ and ${B_Y}={0}_{m_y\times \ell}$. Then, it processes the $i$-th column of $X$ and $Y$ as follows
\begin{equation*}
\begin{split}
&\text{Insert }{X}_i\text{ into a zero valued column of }{B_X}\\
&\text{Insert }{Y}_i\text{ into a zero valued column of }{B_Y}\\
&\textbf{if } {B_X,B_Y}\text{ have no zero valued column} \textbf{ then}\\
&\quad\quad B_X,B_Y = \dcod(B_X,B_Y)\\
&\textbf{end if}
\end{split}
\end{equation*}
for $i=1,\cdots,n$. Each time $B_X,B_Y$ have no zero valued column, performing $B_X,B_Y = \dcod(B_X,B_Y)$ can ensure that the last $1+\ell/2$ columns of $B_X$ and $B_Y$ are zero valued. Therefore, it is easy to verify that the space complexity of COD is only $O((m_x+m_y)\ell)$. However, it needs to compute the QR decomposition of ${B}_X,B_Y$ and SVD of $R_XR_Y^T$ almost $O(n/\ell)$ times, which implies that its time complexity is \[O\left(\frac{n}{\ell}(m_x\ell^2+m_y\ell^2+\ell^3)\right)=O(n(m_x+m_y+\ell)\ell).\]
We will reduce the time complexity by utilizing the sparsity of the input matrices. Our key idea is to employ simultaneous iteration to reduce the number of QR decompositions and SVD.
\begin{algorithm}[t]
\caption{Dense Shrinkage (DS)}
\label{DenseCoD}
\begin{algorithmic}[1]
\STATE \textbf{Input:} $B_X\in\R^{m_x\times \ell^\prime},B_Y\in\R^{m_y\times \ell^\prime}$
\STATE $Q_X,R_X= \qr(B_X)$
\STATE $Q_Y,R_Y= \qr(B_Y)$
\STATE $U,\Sigma,V= \svd(R_XR_Y^\top)$
\STATE $\gamma=\sigma_{\ell^\prime/2}(\Sigma)$
\STATE $\widetilde{\Sigma}=\max(\Sigma-\gamma I_{\ell^\prime},0)$
\STATE $B_X= Q_XU\sqrt{\widetilde{\Sigma}}$
\STATE $B_Y= Q_YV\sqrt{\widetilde{\Sigma}}$
\STATE \textbf{return} $B_X,B_Y$
\end{algorithmic}
\end{algorithm}
\subsection{Simultaneous Iteration}
Simultaneous iteration \cite{SIPower15} is a randomized method for approximate SVD, which provides a very efficient way to handle sparse matrices. Specifically, given a matrix $A\in\mathbb{R}^{m_x\times m_y}$ and  an error coefficient $\epsilon$, it performs the following procedures
\begin{equation}
\label{si_eq1}
\begin{split}
&q=O\left(\log(m_x)/\epsilon\right),G\sim\mathcal{N}(0,1)^{m_y\times \ell}\\
&K=(AA^T)^qAG\\
&\text{Orthonormalize the columns of } K \text{ to obtain } Q
\end{split}
\end{equation}
to generate an orthonormal matrix $Q\in\mathbb{R}^{m_x\times \ell}$, where $G\sim\mathcal{N}(0,1)^{m_y\times \ell}$ denotes that each entry of $G$ is independently sampled from the standard Gaussian distribution $\mathcal{N}(0,1)$. SI enjoys the the following guarantee (\citeauthor{SIPower15}, \citeyear{SIPower15}, Theorem 10).
\begin{thm}
\label{thm_SI}
With probability $99/100$, applying (5) to any matrix $A\in\mathbb{R}^{m_x\times m_y}$ has
\[\|A-QQ^TA\|\leq(1+\epsilon)\sigma_{\ell+1}(A).\]
\end{thm}
Note that some earlier studies \cite{Rokhlin09,Halko11,Woodruff2014,Witten14} have also analyzed this algorithm and achieved similar results.

We will utilize SI to approximately decompose $A=S_XS_Y^T$, where $S_X\in\R^{m_x\times d},S_Y\in\R^{m_y\times d}$ are two buffer matrices, which will be used to store non-zero entries and satisfy $d\leq m=\max(m_x,m_y)$, $\nnz(S_X)\leq m\ell$ and $\nnz(S_Y)\leq m\ell$. The detailed procedures are derived by substituting $A=S_XS_Y^T$ into (\ref{si_eq1}), and are summarized in Algorithm \ref{SI}. 
Specifically, lines 3 to 6 in SI is designed to compute \[K=(S_XS_Y^TS_YS_{X}^T)^qS_XS_Y^TG\] in
$O((\nnz(S_X)+\nnz(S_Y))\ell\log(m_x))$ time, which requires $O((m_x+m_y+d)\ell)$ space. Line 7 in SI can be implemented by Gram-Schmidt orthogonalization or Householder reflections, which only requires $O(m_x\ell^2)$ time and $O(m_x\ell)$ space. Due to $d\leq m=\max(m_x,m_y)$, the space complexity of SI is only $O(m\ell)$, and the time complexity of SI is
\begin{equation}
\label{time_SI}
O((\nnz(S_X)+\nnz(S_Y))\ell\log(m_x)+m_x\ell^2)
\end{equation}
which is efficient for sparse $S_X$ and $S_Y$. By comparison, decomposing  $S_XS_Y^T$ with DS requires $O((m_x+m_y)d^2+d^3)$ time and $O(md)$ space, which is unacceptable for large $d$ even if $S_X$ and $S_Y$ are very sparse.
\begin{algorithm}[t]
\caption{Simultaneous Iteration (SI)}
\label{SI}
\begin{algorithmic}[1]
\STATE \textbf{Input:} $S_X\in\R^{m_x\times d},S_Y\in\R^{m_y\times d},\ell,0<\epsilon<1$
\STATE $q=O\left(\log(m_x)/\epsilon\right)$, $G\sim\mathcal{N}(0,1)^{m_y\times \ell}$
\STATE $K=S_X(S_Y^TG)$
\WHILE{$q>0$}
\STATE $K=S_X(S_Y^T(S_Y(S_{X}^TK))),q=q-1$
\ENDWHILE
\STATE Orthonormalize the columns of $K$ to obtain $Q$
\STATE \textbf{return} $Q,S_Y(S_X^TQ)$
\end{algorithmic}
\end{algorithm}
\section{Main Results}
In this section, we first incorporate simultaneous iteration with a verification process, which is necessary for controlling the failure probability of our algorithm. Then, we describe our sparse co-occuring directions for AMM with limited space and its theoretical results. Finally, we provide a detailed space and runtime analysis of our algorithm. The proofs for theoretical results can be found in the appendix.
\subsection{Verified Simultaneous Iteration}
From previous discussions, in the simple case $n\leq m=\max(m_x,m_y)$, $\nnz(X)\leq m\ell$ and $\nnz(Y)\leq m\ell$, we can generate $B_X$ and $B_Y$ by performing
\[B_X,B_Y=\si(X,Y,\ell,1/10).\]
According to Theorem \ref{thm_SI}, with probability $99/100$
\begin{equation*}
\begin{split}
\|XY^T-B_XB_Y^T\|
\leq&\left(1+\frac{1}{10}\right)\sigma_{\ell+1}(XY^T).
\end{split}
\end{equation*}
Although $X$ and $Y$ generally have more non-zero entries and columns, we could divide $X$ and $Y$ into several smaller matrices that satisfy the conditions of the above simple case, and repeatedly perform SI. However, in this way, the failure probability will increase linearly, where failure means that there exists a run of SI, after which the error between its input and output is unbounded. To reduce the failure probability, we need to verify whether the error between the input and output is bounded by a small value after each run of SI.

\citet{SFD16} have proposed an algorithm to verify the spectral norm of a symmetric matrix. Inspired by their algorithm, we propose verified simultaneous iteration (VSI) as described in Algorithm \ref{BoostedSI}, where $\delta$ is the failure probability. Let $S_X$ and $S_Y$ be its two input matrices. In line 3 of VSI, we use $j$ to record the number of invocations of VSI and set $p=\left\lceil\log(2j^2\sqrt{m_xe}/\delta)\right\rceil$, where $e$ is Euler's number. In line 4 of VSI, we set $\Delta=\frac{11}{10\ell}\sum_{i=1}^{\cols(S_X)}\|S_{X,i}\|_2\|S_{Y,i}\|_2$, where $\cols(S_X)$ denotes the column number of $S_X$. From lines 5 to 12 of VSI, we first utilize SI to generate $C_X,C_Y$, and then verify whether
\begin{equation}
\label{verify_cond}
\|(CC^T)^{p}\mathbf{x}\|_2\leq \|\mathbf{x}\|_2
\end{equation}
holds, where $C=(S_X S_Y^{T}-C_XC^{T}_Y)/\Delta$ and $\mathbf{x}$ is drawn from $\mathcal{N}(0,1)^{m_x\times 1}$. If so, we will return $C_X,C_Y$. Otherwise, we will rerun SI and repeat the verification process until it holds.

Note that the condition (\ref{verify_cond}) is used to verify whether $\|C\|>2$, and if $C$ satisfies this condition, with high probability, $\|C\|\leq2$. Specifically, we establish the following guarantee.
\begin{lem}
\label{lem2}
Assume that $C_X,C_Y$ are returned by the $j$-th run of VSI, with probability at least $1-\frac{\delta}{2j^2}$,
\[\left\|S_X S_Y^{T}-C_XC^{T}_Y\right\|\leq2\Delta\]
where $\Delta=\frac{11}{10\ell}\sum_{i=1}^{\cols(S_X)}\|S_{X,i}\|_2\|S_{Y,i}\|_2$.
\end{lem}
Lemma \ref{lem2} implies that the failure probability of bounding $\left\|S_X S_Y^{T}-C_XC^{T}_Y\right\|$ decreases as the number of invocations of VSI increases, instead of keeping $1/100$ for the naive SI, which is essential for our analysis.
\begin{algorithm}[t]
\caption{Verified Simultaneous Iteration (VSI)}
\label{BoostedSI}
\begin{algorithmic}[1]
\STATE \textbf{Input:} $S_X\in\R^{m_x\times d},S_Y\in\R^{m_y\times d},\ell,0<\delta<1$
\STATE \textbf{Initialization:} persistent $j=0$ ($j$ retains its value between invocations)
\STATE $j=j+1$, $p=\left\lceil\log(2j^2\sqrt{m_xe}/\delta)\right\rceil$
\STATE $\Delta=\frac{11}{10\ell}\sum_{i=1}^{\cols(S_X)}\|S_{X,i}\|_2\|S_{Y,i}\|_2$
\WHILE{True}
\STATE $C_X,C_Y=\si(S_X,S_Y,\ell,1/10)$
\STATE $C=(S_X S_Y^{T}-C_XC^{T}_Y)/\Delta$ ($C$ is not computed)
\STATE $\mathbf{x}\sim\mathcal{N}(0,1)^{m_x\times 1}$
\IF{$\|(CC^T)^{p}\mathbf{x}\|_2\leq \|\mathbf{x}\|_2$}
\STATE \textbf{return} $C_X,C_Y$
\ENDIF
\ENDWHILE
\end{algorithmic}
\end{algorithm}
\subsection{Sparse Co-occuring Directions}
To work with limited space and exploit the sparsity, we propose an efficient variant of COD for sparse matrices as follows.

Let ${X}\in\R^{m_x\times n}$ and ${Y}\in\R^{m_y\times n}$ be the two input matrices. In the beginning, we initialize ${B_X}={0}_{m_x\times \ell}$ and ${B_Y}={0}_{m_y\times \ell}$, where $\ell\ll\min(m_x,m_y,n)$. Moreover, we initialize two empty buffer matrices as $S_{X}={0}_{m_x\times 0}$ and $S_{Y}={0}_{m_y\times 0}$, which will be used to store the non-zero entries of $X$ and $Y$. %
To avoid excessive space cost, the buffer matrices are deemed full when $S_X$ or $S_Y$ contains $m\ell$ non-zero entries or $m$ columns. For $i=1,\cdots,n$, after receiving $X_i$ and $Y_i$, we store the non-zero entries of them in the buffer matrices as
\[S_{X}=[S_{X},{X}_i],S_{Y}=[S_{Y},Y_i]\]
where $[\cdot,\cdot]$ concatenates two matrices horizontally. Each time the buffer matrices become full, we first utilize VSI in Algorithm \ref{BoostedSI} to approximately decompose $S_XS_Y^T$ as
\[C_X,C_Y=\bsi(S_{X},S_{Y},\ell,\delta)\]
where $C_X\in\mathbb{R}^{m_x\times\ell},C_Y\in\mathbb{R}^{m_y\times\ell}$ and $\delta$ is the failure probability. Then, to merge $C_X,C_Y$ into $B_X,B_Y$, we utilize DS in Algorithm \ref{DenseCoD} as follows
\begin{align*}
&D_X=[B_X,C_X],D_Y=[B_Y,C_Y]\\
&B_X,B_Y=\dcod(D_X,D_Y)
\end{align*}
where $B_X\in\mathbb{R}^{m_x\times\ell},B_Y\in\mathbb{R}^{m_y\times\ell}$ are large enough to store the non-zero valued columns returned by DS. Finally, we reset the buffer matrices as
\[S_{X}={0}_{m_x\times 0},S_{Y}={0}_{m_y\times 0}\]
and continue to process the remaining columns in the same way. The detailed procedures of our algorithm are summarized in Algorithm \ref{SparseCoD} and it is named as sparse co-occuring directions (SCOD).

\begin{algorithm}[t]
\caption{Sparse Co-occuring Directions (SCOD)}
\label{SparseCoD}
\begin{algorithmic}[1]
\STATE \textbf{Input:} $X\in\R^{m_x\times n},Y\in\R^{m_y\times n},\ell,0<\delta<1$
\STATE \textbf{Initialization:} $B_X={0}_{m_x\times l},B_X={0}_{m_y\times l}$, $S_{X}={0}_{m_x\times 0},S_{Y}={0}_{m_y\times 0}$
\FOR{$i=1,...,n$}
\STATE $S_{X}=[S_{X},{X}_i],S_{Y}=[S_{Y},Y_i]$
\IF{$\nnz(S_{X})\geq \ell m$ \textbf{or} $\nnz(S_{Y})\geq \ell m$ \textbf{or} $\cols(S_{X})=m$}
\STATE $C_X,C_Y=\bsi(S_{X},S_{Y},\ell,\delta)$
\STATE $D_X=[B_X,C_X],D_Y=[B_Y,C_Y]$
\STATE $B_X,B_Y=\dcod(D_X,D_Y)$
\STATE $S_{X}={0}_{m_x\times 0},S_{Y}={0}_{m_y\times 0}$
\ENDIF
\ENDFOR
\STATE \textbf{return} $B_X,B_Y$
\end{algorithmic}
\end{algorithm}
Following \citet{CoD17}, we first bound the approximation error of our SCOD as below.
\begin{thm}
\label{thm1}
Given $X\in\mathbb{R}^{m_x\times n},Y\in\mathbb{R}^{m_y\times n},\ell\leq\min(m_x,m_y,n)$ and $\delta\in(0,1)$, Algorithm \ref{SparseCoD} outputs $B_X\in\mathbb{R}^{m_y\times \ell},B_Y\in\mathbb{R}^{m_y\times \ell}$ such that
\begin{align*}
\|XY^T-B_XB_Y^T\|\leq\frac{16\|X\|_F\|Y\|_F}{5\ell}
\end{align*}
with probability at least $1-\delta$.
\end{thm}
Compared with the error bound (\ref{S_bound2}) of COD, the error bound of our SCOD only magnifies it by a small constant factor of $1.6$. Furthermore, the following theorem shows that the output of SCOD can be used to compute a rank-$k$ approximation for $XY^T$.
\begin{thm}
\label{thm2}
Let $B_X\in\mathbb{R}^{m_y\times \ell},B_Y\in\mathbb{R}^{m_y\times \ell}$ be the output of Algorithm \ref{SparseCoD}. Let $k\leq \ell$ and $\bar{U},\bar{V}$ be the matrices whose columns are the top-$k$ left and right singular vectors of $B_XB_Y^T$. Let $\pi_{\bar{U}(X)}=\bar{U}\bar{U}^TX$ and $\pi_{\bar{V}}(Y)=\bar{V}\bar{V}^TY$. For $\epsilon>0$ and $\ell\geq\frac{64\sqrt{\sr(X)\sr(Y)}}{5\epsilon}\frac{\|X\|\|Y\|}{\sigma_{k+1}(XY^T)}$, we have
\begin{align}
\label{k-error}
\|XY^T-\pi_{\bar{U}}(X)\pi_{\bar{V}}(Y)^T\|\leq\sigma_{k+1}(XY^T)(1+\epsilon)
\end{align}
with probability at least $1-\delta$.
\end{thm}
\citet{CoD17} have proved that the output of COD enjoys (\ref{k-error}) with \begin{equation}
\label{cod_l}
\ell\geq\frac{8\sqrt{\sr(X)\sr(Y)}}{\epsilon}\frac{\|X\|\|Y\|}{\sigma_{k+1}(XY^T)}.
\end{equation} By contrast, the lower bound of $\ell$ for our SCOD only magnifies the right term in (\ref{cod_l}) by a small constant factor of $1.6$.
\subsection{Space and Runtime Analysis}
The total space complexity of our SCOD is only $O(m\ell)$, as explained below.
\begin{itemize}
\item $B_X,B_Y,S_X,S_Y,C_X,C_Y,D_X,D_Y$ maintained in Algorithm \ref{SparseCoD} only require $O(m\ell)$ space.
\item Because of the \emph{if} conditions in SCOD, VSI invoked by Algorithm \ref{SparseCoD} only requires $O(m\ell)$ space.
\item Because of $\cols(D_X)=\cols(D_Y)=2\ell$, DS invoked by Algorithm \ref{SparseCoD} only requires $O(m\ell)$ space.
\end{itemize}
To analyze the expected runtime of SCOD, we first note that it is dominated by the cumulative runtime of VSI and DS that are invoked after each time the \emph{if} statement in Algorithm \ref{SparseCoD} is triggered. Without loss of generality, we assume that the \emph{if} statement is triggered $s$ times in total. It is not hard to verify \begin{equation}
\label{s_value}s\leq\frac{\nnz(X)+\nnz(Y)}{m\ell}+\frac{n}{m}.\end{equation}
Because of $\cols(D_X)=\cols(D_Y)=2\ell$, each run of DS requires $O(m\ell^2)$ time. Therefore, the total time spent by invoking DS $s$ times is
\begin{equation}
\label{time_DS}
O\left(sm\ell^2\right)=O\left((\nnz(X)+\nnz(Y))\ell+n\ell^2\right).
\end{equation}
Then, we further bound the expected cumulative runtime of VSI. It is not hard to verify that the runtime of VSI is dominated by the time spent by lines 6 and 9 in Algorithm \ref{BoostedSI}. According to Theorem \ref{thm_SI}, after each execution of line 6 in Algorithm \ref{BoostedSI}, with probability $99/100$, we have
\begin{align*}
\|S_X S_Y^{T}-C_XC^{T}_Y\|&\leq(1+\frac{1}{10})\sigma_{\ell+1}\left(S_XS_Y^{T}\right)\\
&\leq\frac{11}{10\ell}\|S_X S_Y^{T}\|_\ast\\
&\leq\frac{11}{10\ell}\sum_{i=1}^{\cols(S_X)}\|S_{X,i}S_{Y,i}^T\|_\ast\\
&\leq\frac{11}{10\ell}\sum_{i=1}^{\cols(S_X)}\|S_{X,i}\|_2\|S_{Y,i}\|_2\\
&=\Delta
\end{align*}
which implies $\left\|(S_X S_Y^{T}-C_XC^{T}_Y)/\Delta\right\|\leq 1$.

Combining with $C=(S_X S_Y^{T}-C_XC^{T}_Y)/\Delta$, we have
\begin{align*}
\|(CC^T)^{p}\mathbf{x}\|_2&\leq\|(CC^T)\|^{p}\|\mathbf{x}\|_2\leq \|\mathbf{x}\|_2
\end{align*}
with probability $99/100$. Therefore, for Algorithm \ref{BoostedSI}, the probability that $C_X$ and $C_Y$ generated by executing its line 6 satisfy the \emph{if} condition in its line 9 is $99/100>1/2$. Hence, in each run of VSI, the expected number of executing lines 6 and 9 is at most 2.

Because of the \emph{if} conditions in SCOD and the time complexity of SI in (\ref{time_SI}), each time executing line 6 in VSI requires \[O((\nnz(S_X)+\nnz(S_Y))\ell\log(m_x)+m_x\ell^2)\]
time. Let $S_X^t,S_Y^t$ denote the values of $S_X,S_Y$ in the $t$-th execution of line $6$ in Algorithm \ref{SparseCoD}, where $t=1,\cdots,s$. Because of $\sum_{t=1}^s\nnz(S_X^t)=\nnz(X)$ and $\sum_{t=1}^s\nnz(S_Y^t)=\nnz(Y)$, in expectation, the total time spent by executing line 6 in VSI is \begin{equation}
\label{BSS_SS_time}
O((\nnz(X)+\nnz(Y))\ell\log(m_x)+n\ell^2).
\end{equation}
In the $j$-th run of VSI, each time executing its line 9 needs to compute $\|(CC^T)^{p}\mathbf{x}\|_2$, which requires $O\left(m\ell p\right)$ time, because $C=(S_X S_Y^{T}-C_XC^{T}_Y)/\Delta$ and $S_X,S_Y,C_X,C_Y$ have less than $O(m\ell)$ non-zero entries. Note that $p=\left\lceil\log(2j^2\sqrt{m_xe}/\delta)\right\rceil$. So, in expectation, the total time spent by executing line 9 in VSI is
\begin{align*}
\sum_{j=1}^{s}O\left(m\ell\log(m_xj/\delta)\right)\leq O\left(sm\ell\log(m_xs/\delta)\right).
\end{align*}
Finally, combining the above inequality, (\ref{s_value}), (\ref{time_DS}) and (\ref{BSS_SS_time}), the expected runtime of SCOD is
\begin{align*}
O\left(N\ell\log(m_x)+n\ell^2+(N+n\ell)\log(n/\delta)\right)
\end{align*}
where $N=\nnz(X)+\nnz(Y)$.
\subsection{Discussions}
First, we note that there exist numerical softwares (e.g., Matlab), which provide highly optimized subroutine to efficiently compute the exact multiplication $XY^T$, if $X$ and $Y$ are sparse. However, they suffer a high space complexity of $\nnz(X)+\nnz(Y)+\nnz(XY^T)$ to operate $X$, $Y$ and $XY^T$ in the main memory, which is not applicable for large matrices when the memory space is limited. By contrast, the space complexity of our SCOD is only $O(m \ell)$.

Second, when this paper is under review, we find that a concurrent work \citep{Luo2020} independently proposed almost the same algorithm as our work. The only difference between the algorithm of \citet{Luo2020} and our SCOD is the way to control the failure probability caused by repeatedly invoking SI. In our paper, this problem is addressed by adding a verification process to SI. By contrast, \citet{Luo2020} addressed it by increasing the number of iteration in SI according to the number of invocations of SI. Furthermore, with a more careful analysis, \citet{Luo2020} established the error bound of
\begin{align*}
&\|XY^T-B_XB_Y^T\|\\\leq& O\left(\frac{\|X\|_F\|Y\|_F-\sum_{i=1}^k\sigma_i(XY^T)}{\ell-k}\right)
\end{align*}
for $k<\ell$, which is tighter than our error bound in Theorem \ref{thm1} when $XY^T$ is low-rank. Since their algorithm is almost the same as ours, their error bound actually also provides a better guarantee for our algorithm.

\begin{figure*}[t]
\centering
\subfigure[Approximation Error]{\includegraphics[width=0.33\textwidth]{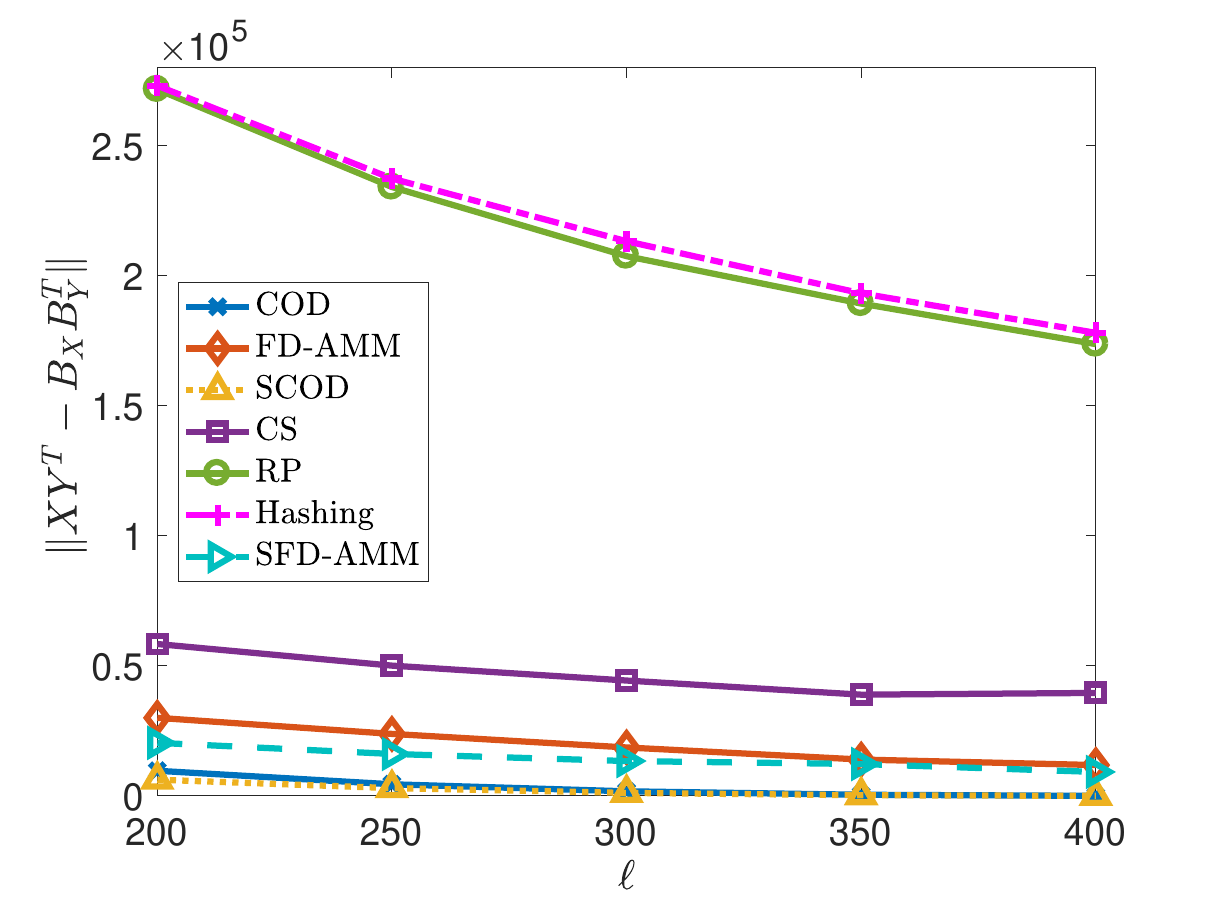}}
\centering
\subfigure[Projection Error]{\includegraphics[width=0.33\textwidth]{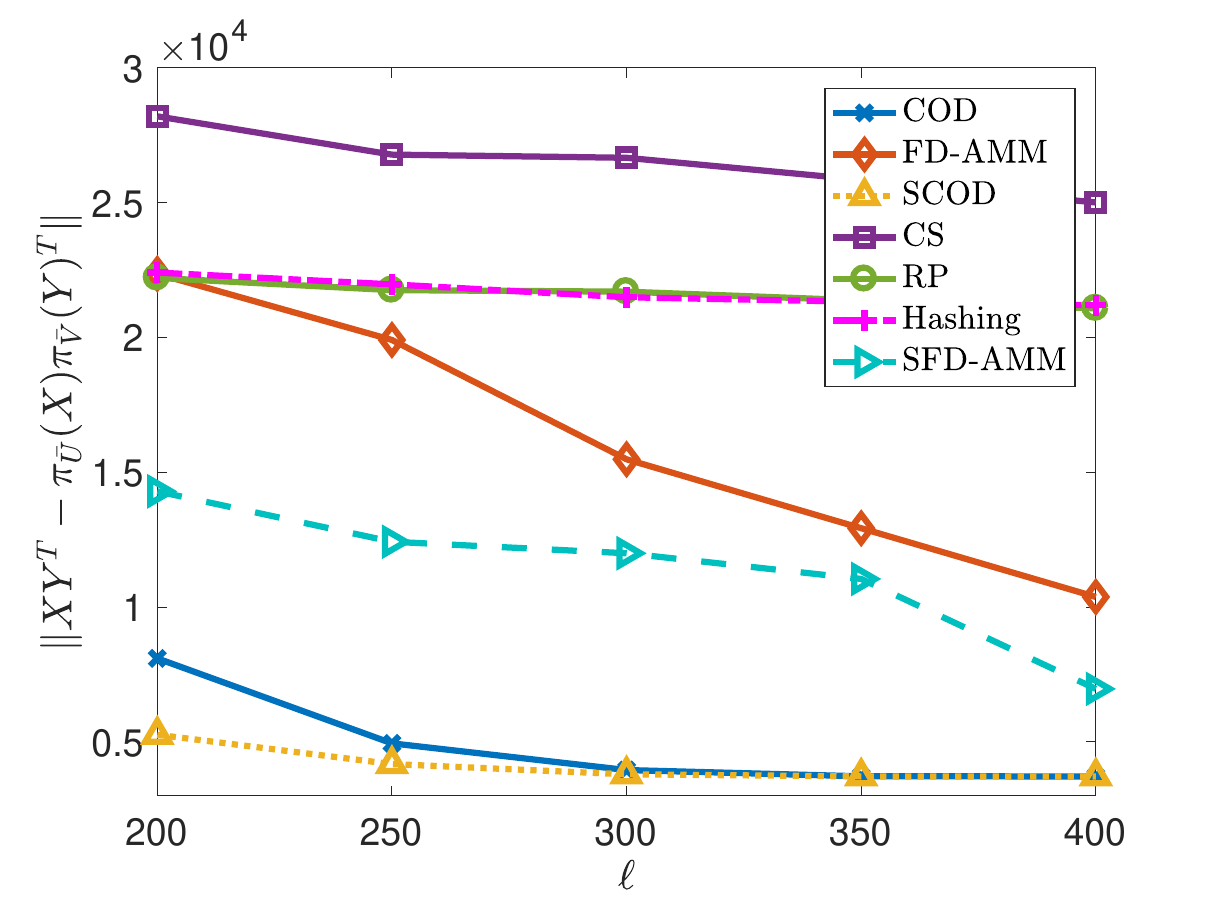}}
\centering
\subfigure[Runtime]{\includegraphics[width=0.33\textwidth]{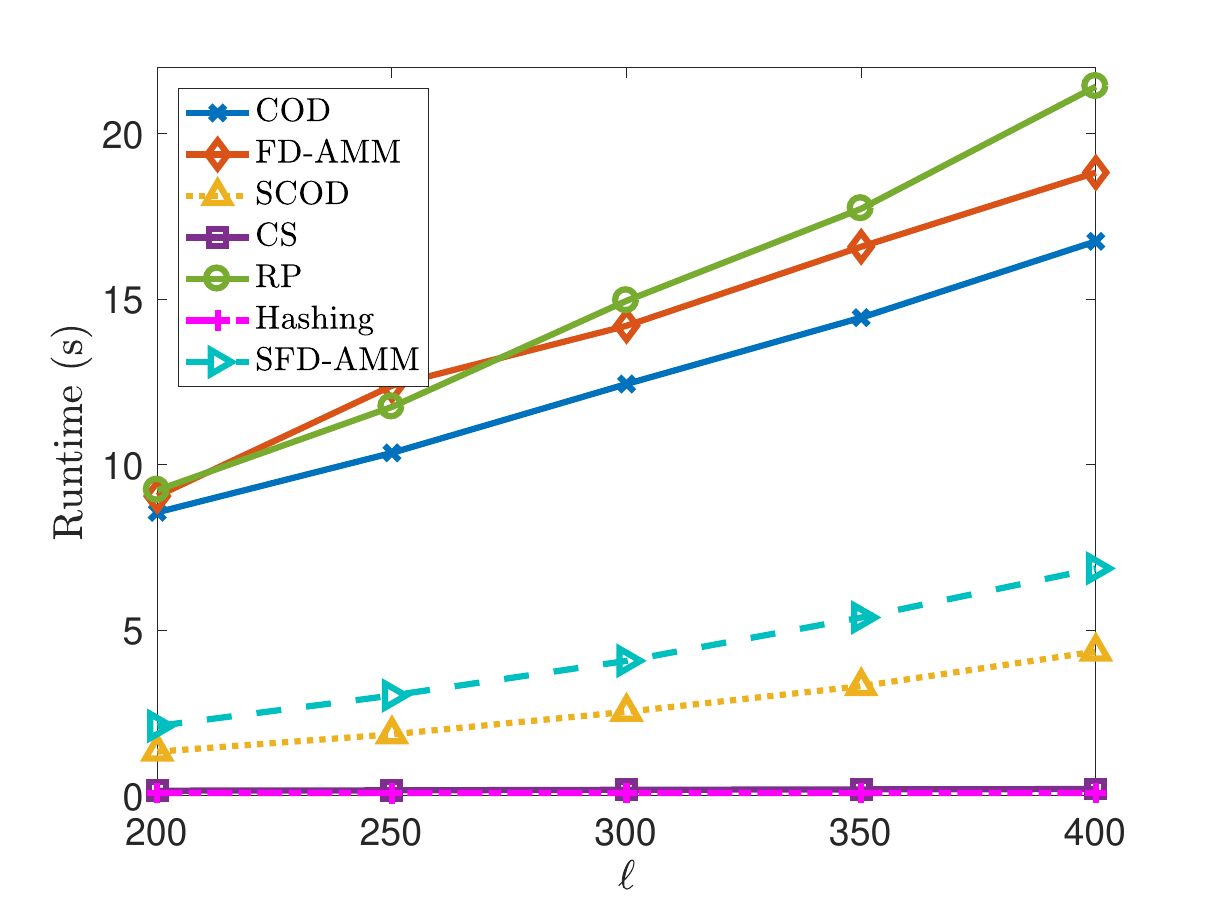}}
\caption{Experimental results among different sketch size on the low-rank dataset.}
\label{fig1}
\centering
\subfigure[Approximation Error]{\includegraphics[width=0.33\textwidth]{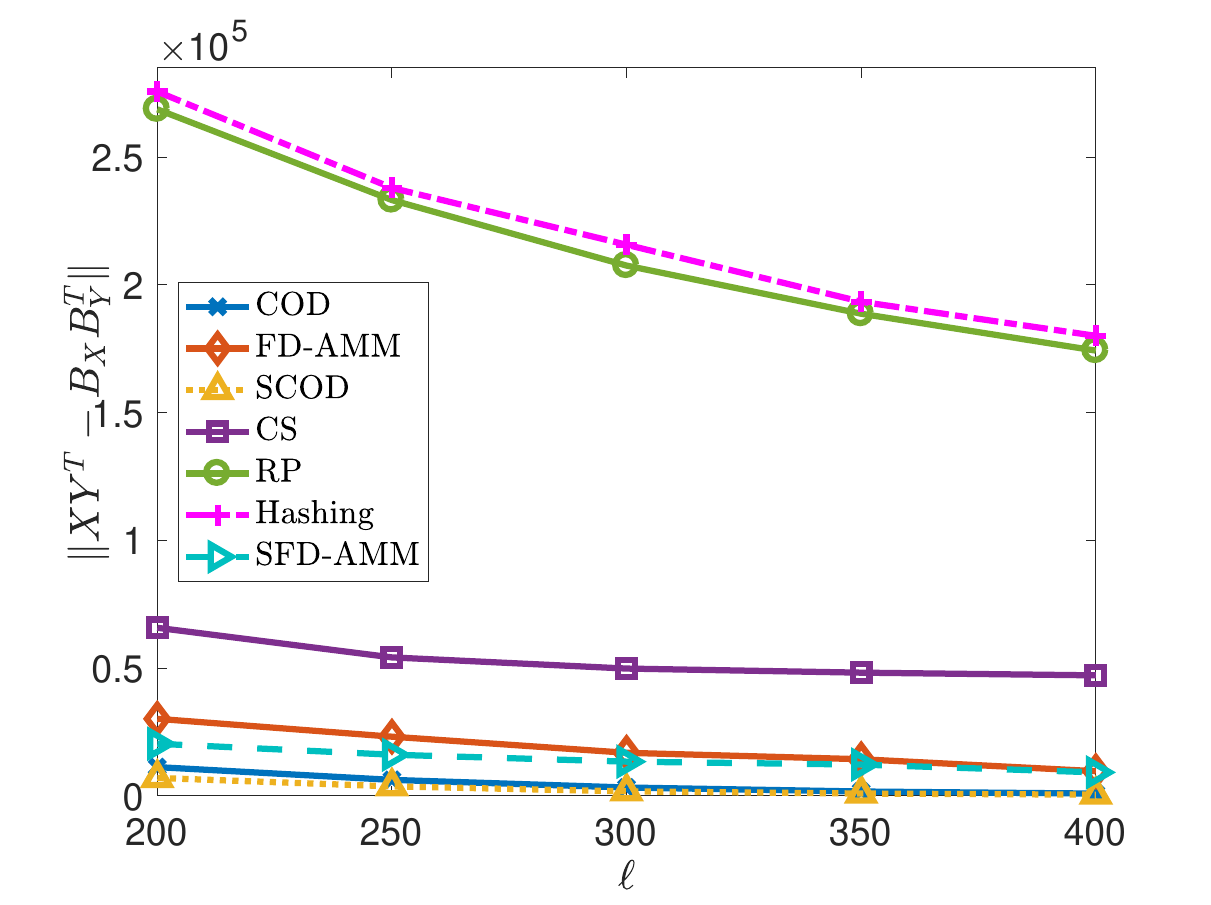}}
\centering
\subfigure[Projection Error]{\includegraphics[width=0.33\textwidth]{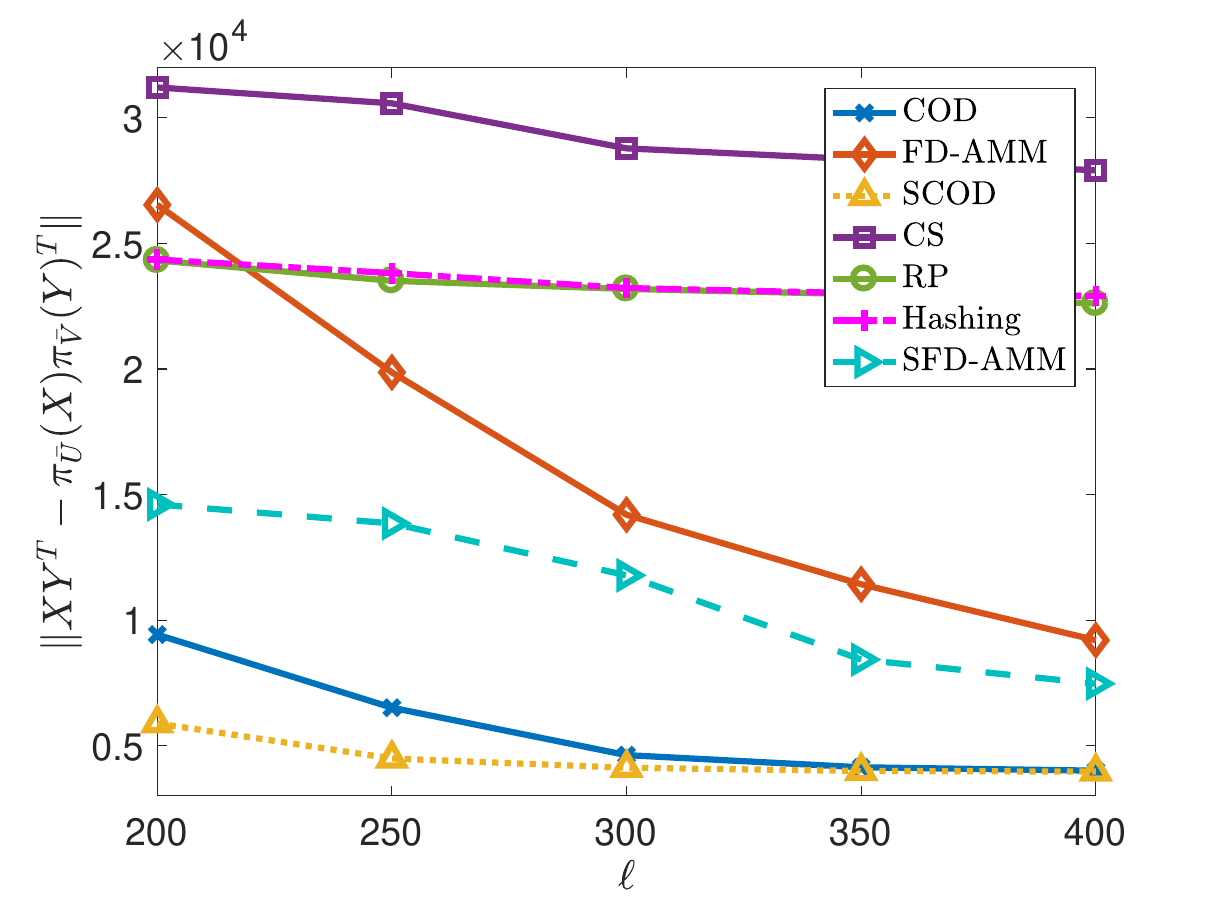}}
\centering
\subfigure[Runtime]{\includegraphics[width=0.33\textwidth]{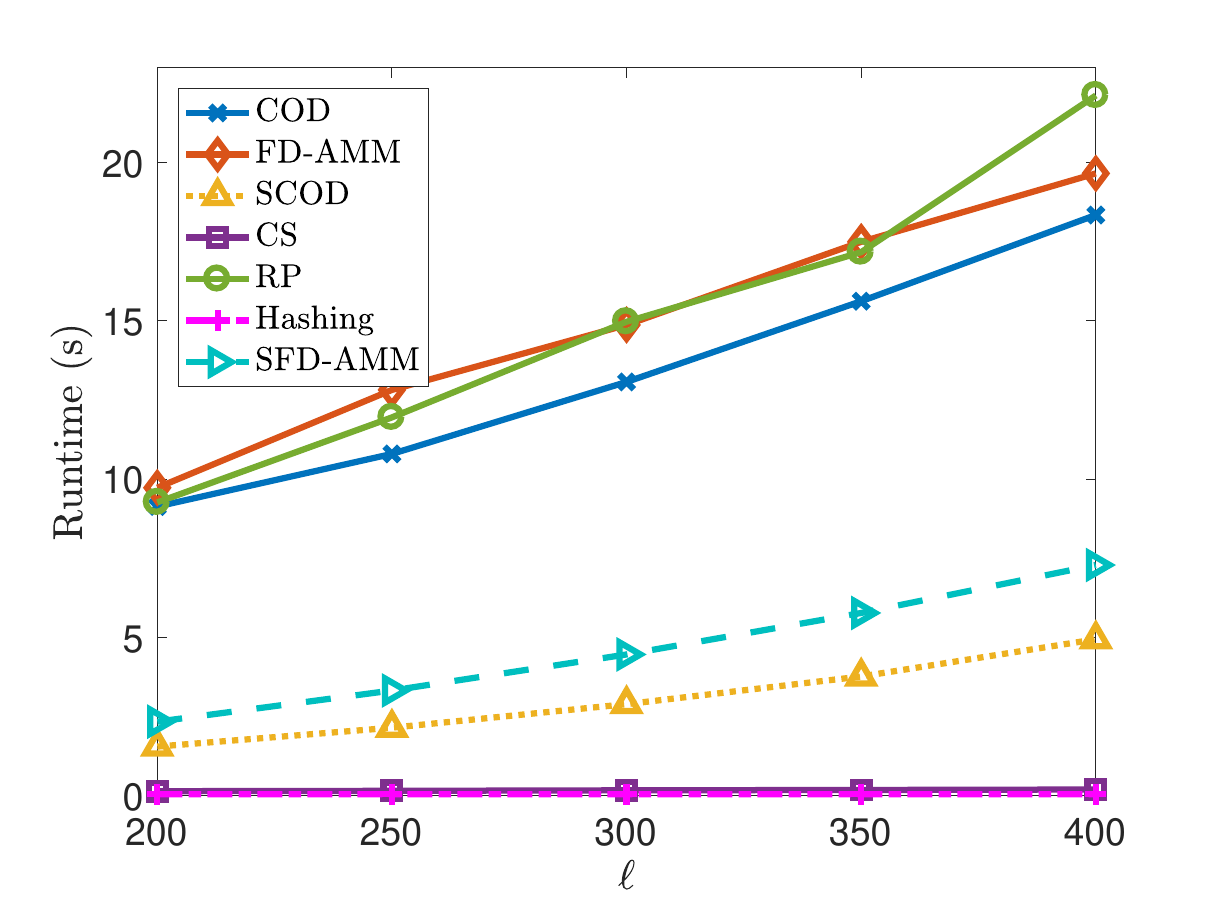}}
\caption{Experimental results among different sketch size on the noisy low-rank dataset.}
\label{fig2}
\end{figure*}

\begin{figure*}[t]
\centering
\subfigure[Approximation Error]{\includegraphics[width=0.33\textwidth]{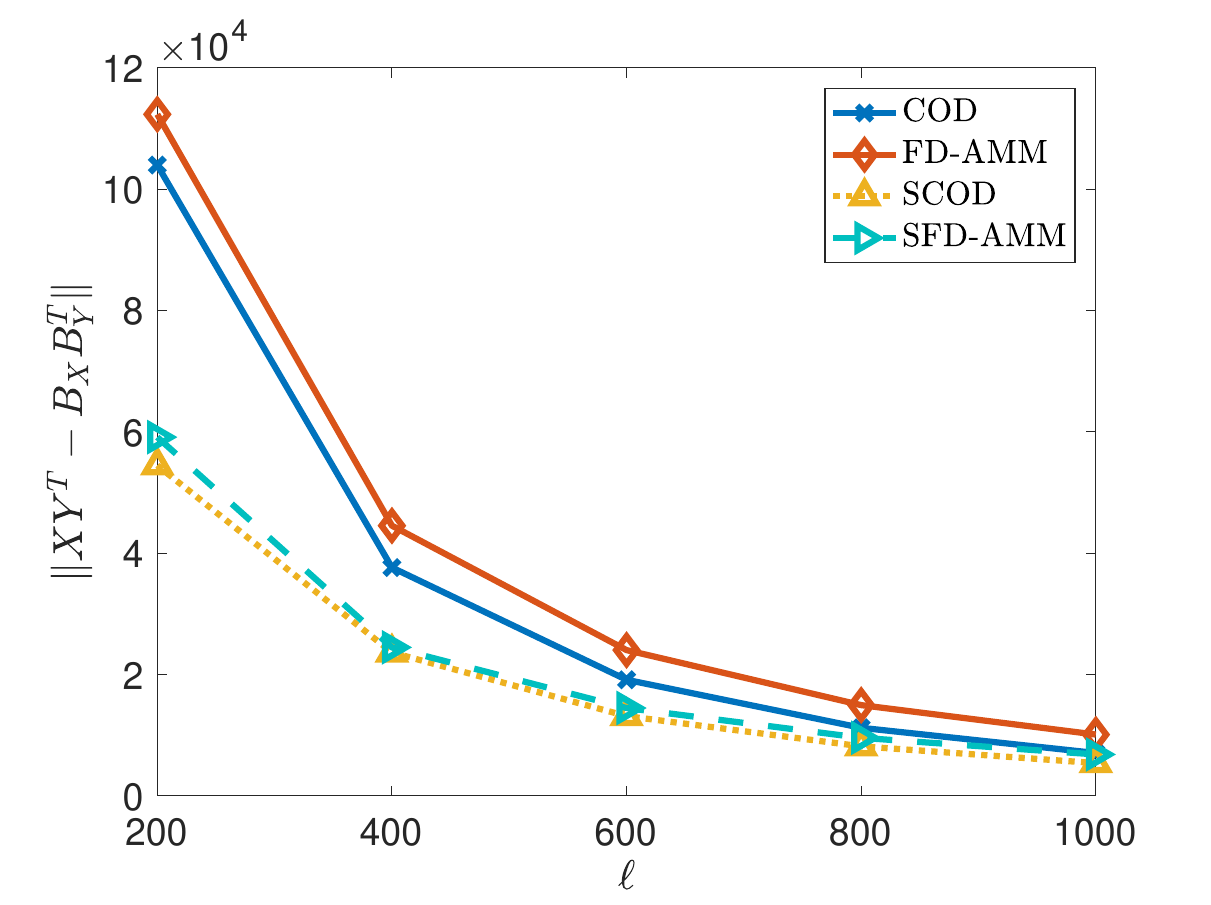}}
\centering
\subfigure[Projection Error]{\includegraphics[width=0.33\textwidth]{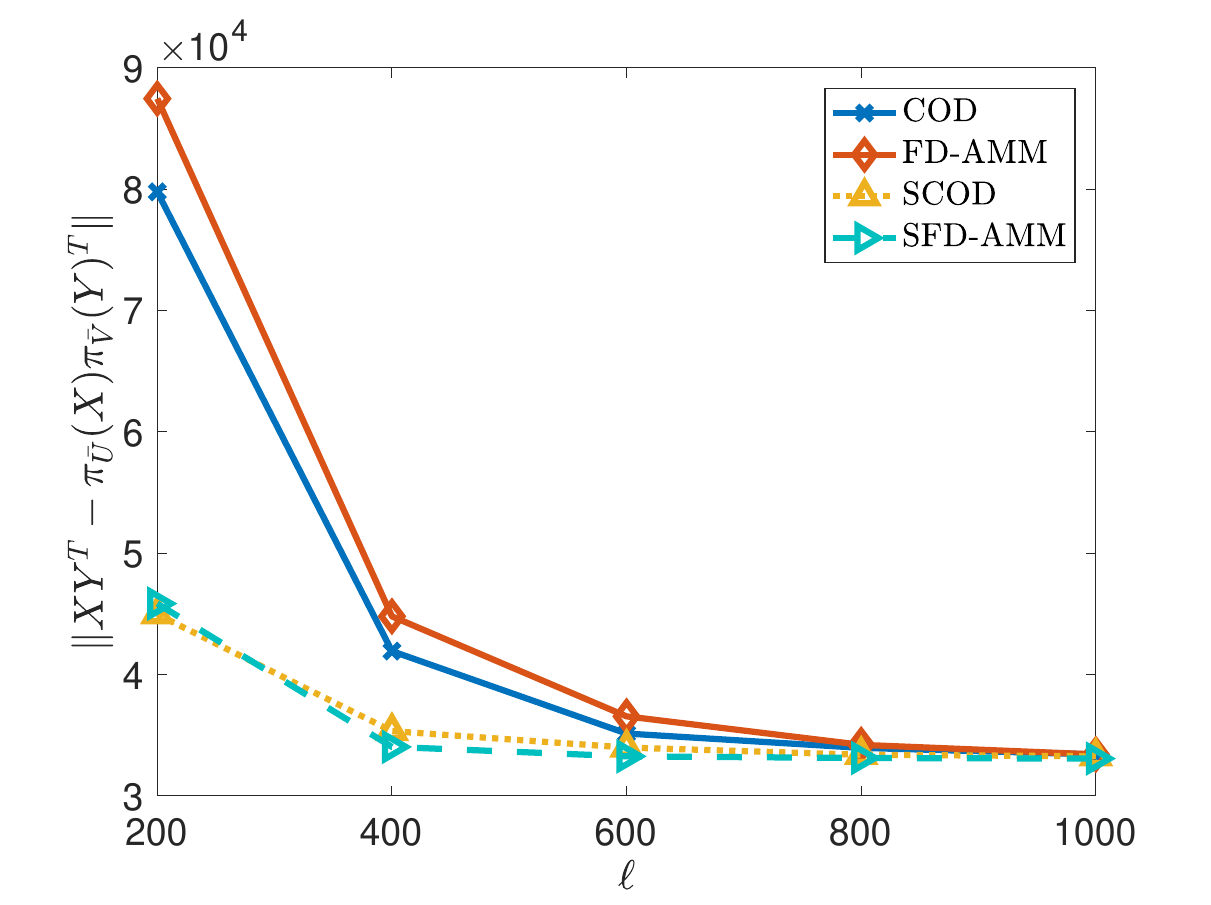}}
\centering
\subfigure[Runtime]{\includegraphics[width=0.33\textwidth]{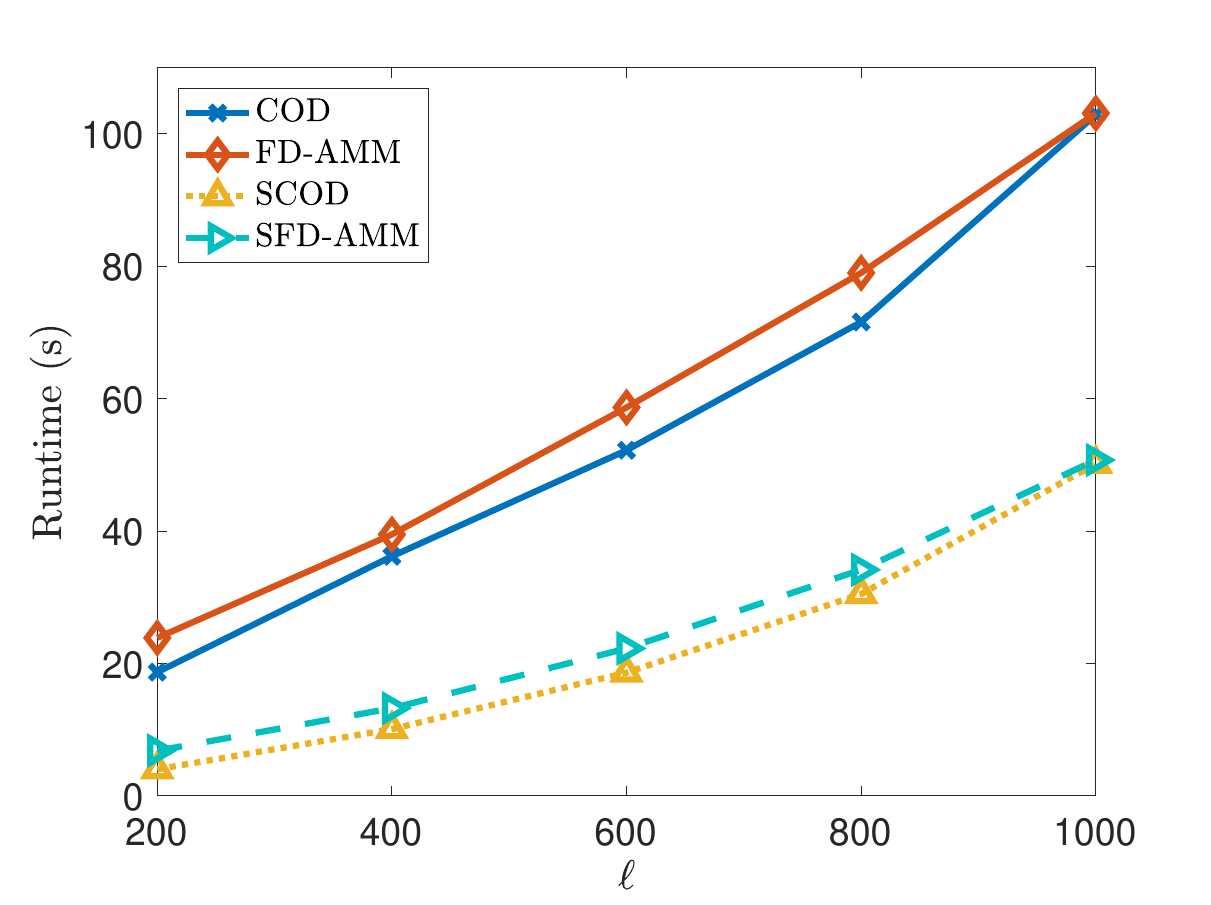}}
\caption{Experimental results among different sketch size on NIPS conference papers dataset.}
\label{fig3}
\centering
\subfigure[Approximation Error]{\includegraphics[width=0.33\textwidth]{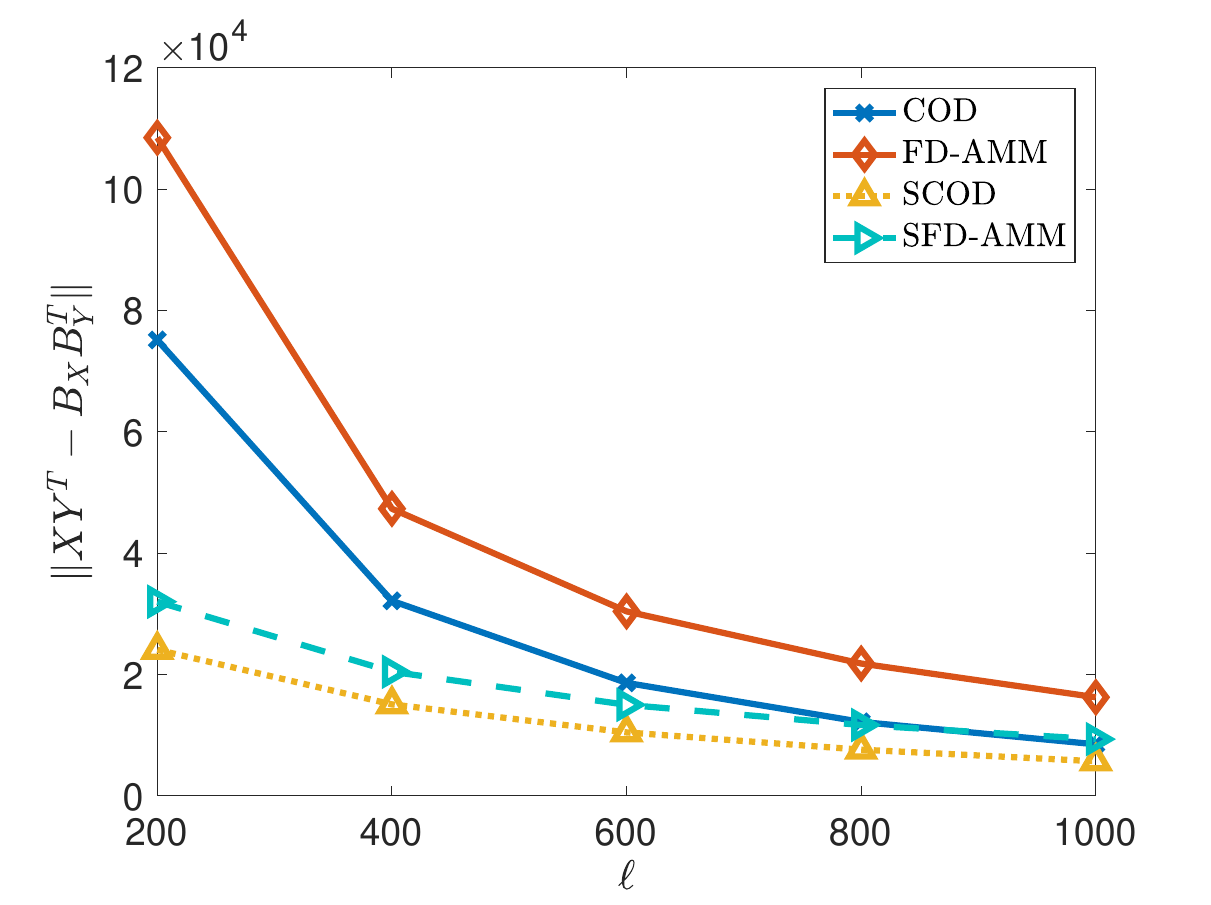}}
\centering
\subfigure[Projection Error]{\includegraphics[width=0.33\textwidth]{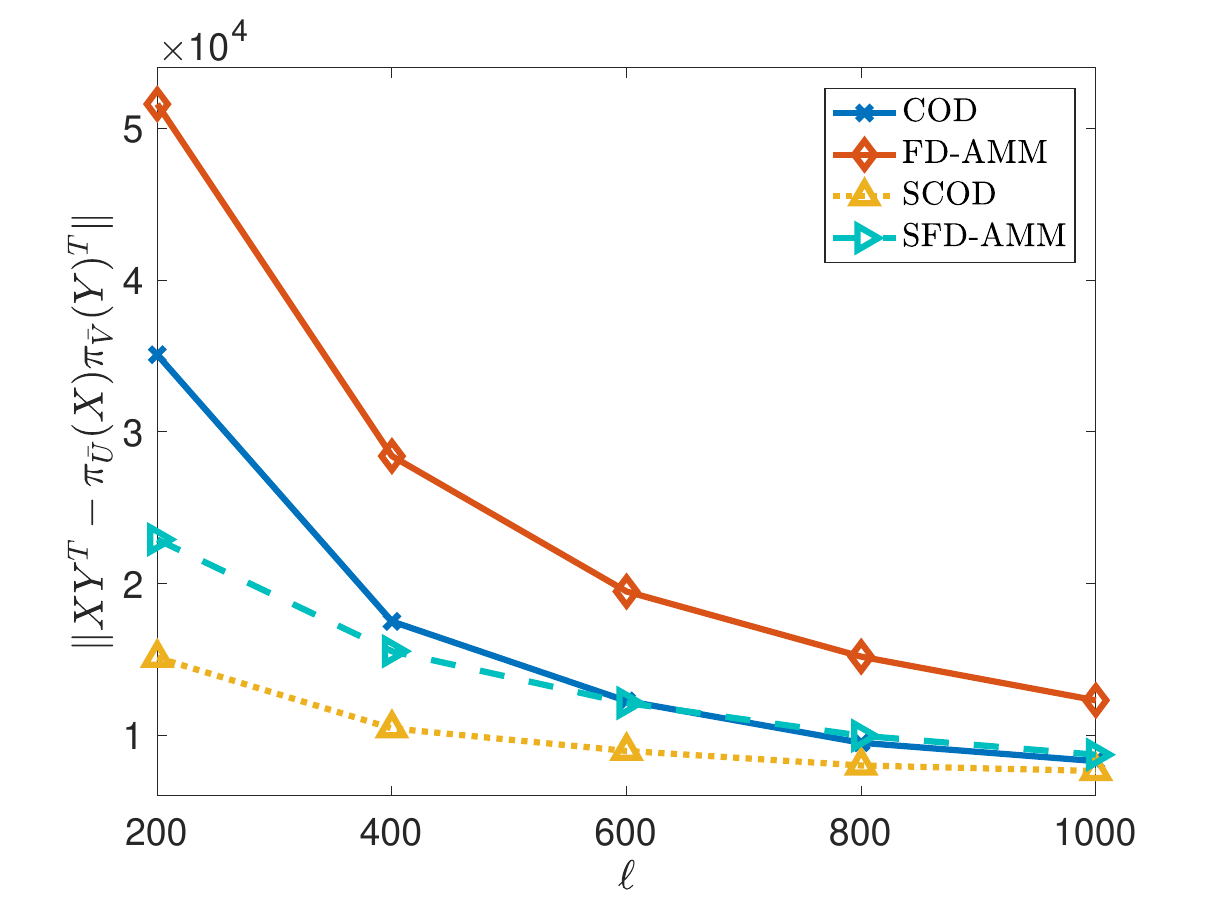}}
\centering
\subfigure[Runtime]{\includegraphics[width=0.33\textwidth]{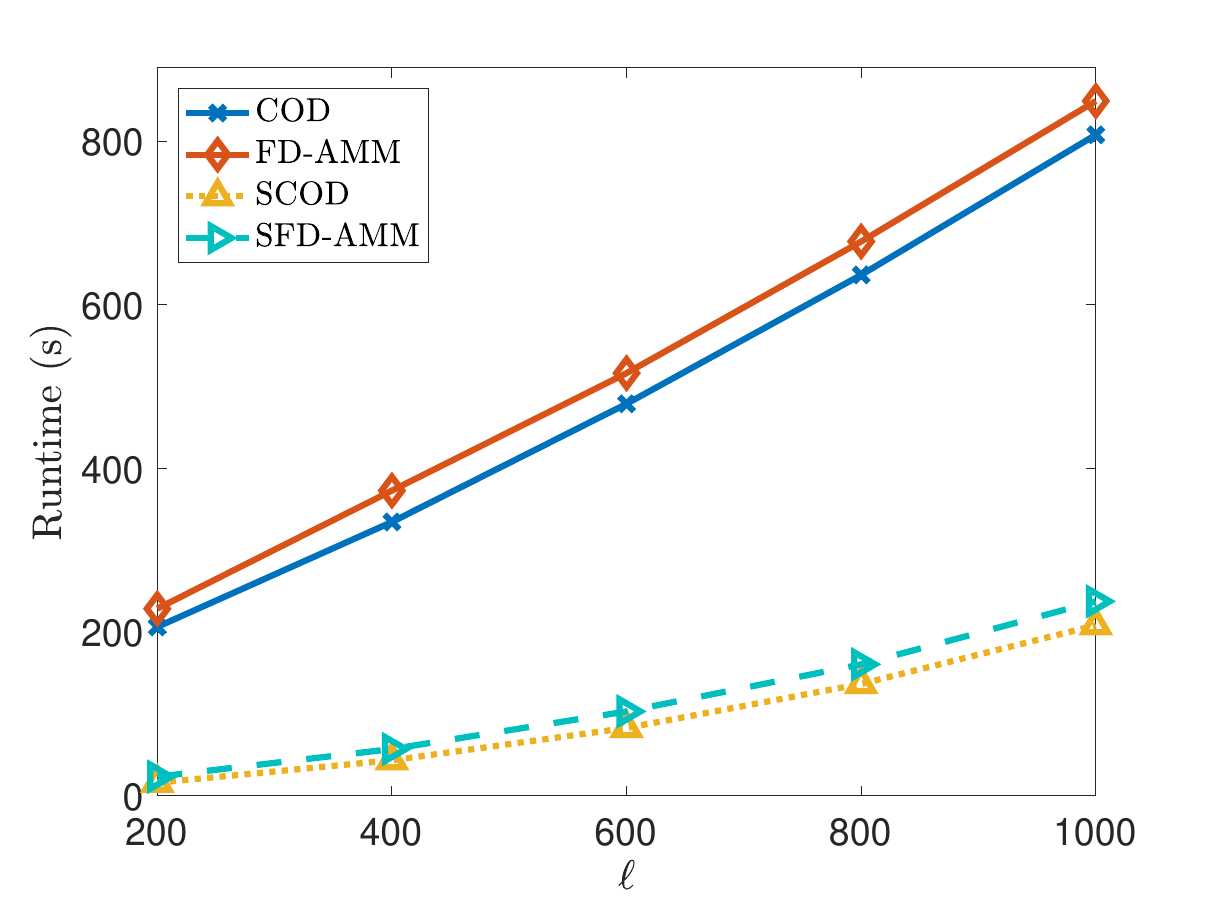}}
\caption{Experimental results among different sketch size on MovieLens 10M dataset.}
\label{fig4}
\end{figure*}
\section{Experiments}
In this section, we perform numerical experiments to verify the efficiency and effectiveness of our SCOD.
\subsection{Datasets}
We conduct experiments on two synthetic datasets and two real datasets: NIPS conference papers\footnote{\url{https://archive.ics.uci.edu/ml/datasets/NIPS+Conference+Papers+1987-2015}} \cite{NIPSpaper} and MovieLens 10M\footnote{\url{https://grouplens.org/datasets/movielens/10m/}}. Each dataset consists of two sparse matrices $X\in\mathbb{R}^{m_x\times n}$ and $Y\in\mathbb{R}^{m_y\times n}$. The synthetic datasets are randomly generated with $\spr$, which is a built-in function of Matlab. Specifically, we first generate a low-rank dataset by setting \begin{align*}
&X=\spr(1e3,1e4,0.01,\mathbf{r})\\
&Y=\spr(2e3,1e4,0.01,\mathbf{r})
\end{align*}
where $\mathbf{r}=[400,399,\cdots,1]\in\mathbb{R}^{400}$, which ensures that $X\in\mathbb{R}^{1e3\times1e4}$ and $Y^{2e3\times1e4}$ only contain $1\%$ non-zero entries and their non-zero singular values are equal to $\mathbf{r}$. With the same $\mathbf{r}$, a noisy low-rank dataset is generated by adding a sparse noise to the above low-rank matrices as
\begin{align*}
&X=\spr(1e3,1e4,0.01,\mathbf{r})+\spr(1e3,1e4,0.01)\\
&Y=\spr(2e3,1e4,0.01,\mathbf{r})+\spr(2e3,1e4,0.01)
\end{align*}
where $X$ and $Y$ contain less than $2\%$ non-zero entries.

Moreover, NIPS conference papers dataset is originally a $11463\times5811$ word-by-document matrix $M$, which contains the distribution of words in 5811 papers published between the years 1987 and 2015. In our experiment, let $X^T$ be the first 2905 columns of $M$, and let $Y^T$ be the others. Therefore, the product $XY^T$ reflects the similarities between two sets of papers. Similarly, MovieLens 10M dataset is originally a $69878 \times 10677$ user-item rating matrix $M$. We also let $X^T$ be the first $5338$ columns of $M$ and $Y^T$ be the others.
\subsection{Baselines and Setting}
We first show that our SCOD can match the accuracy of COD, and significantly reduce the runtime of COD for sparse matrices. Moreover, we compare SCOD against other baselines for AMM with limited space including FD-AMM \cite{Qiaomin16} and the following randomized algorithms:
\begin{itemize}
\item Sparse frequent directions based AMM (SFD-AMM): replacing FD used in FD-AMM with sparse frequent directions (SFD) proposed by \citet{SFD16}, which is suggested by an anonymous reviewer;
\item Column selection (CS) \cite{Drineas06}: independently sampling $\ell$ columns from $X$ and $Y$ with the probability $p_i=\frac{\|X_i\|_2\|Y_i\|_2}{S}$ for each column pair $X_i,Y_i$, and scaling each selected column by $\frac{1}{\sqrt{\ell p_i}}$, where $S=\sum_{i=1}^n\|X_i\|_2\|Y_i\|_2$, which is implemented with $\ell$ independent reservoir samplers for processing the data in a streaming way;
\item Random projection (RP) \cite{Sarlos06}: $B_X=X\Pi$, $B_Y=Y\Pi$, where $\Pi\in\mathbb{R}^{n\times \ell}$ and each entry $\Pi_{ij}$ is uniformly sampled from $\{-1/\sqrt{\ell},1/\sqrt{\ell}\}$;\footnote{We have also tried Gaussian random projection, the results of which are very close to RP. So, those results are omitted.}
\item Hashing \cite{hashing13}: for each $i=1,\cdots,n$, updating $B_X$, $B_Y$ as $B_{X,h(i)}=B_{X,h(i)}+s(i)X_i$, $B_{Y,h(i)}=B_{Y,h(i)}+s(i)Y_i$, where $h(i)$ and $s(i)$ are uniformly sampled from $\{1,\cdots,\ell\}$ and $\{1,-1\}$, respectively.
\end{itemize}
In the previous sections, to control the failure probability of SCOD, we have used VSI in line 6 of Algorithm \ref{SparseCoD}. However, in practice, we find that directly utilizing SI is enough to ensure the accuracy of SCOD. Therefore, in the experiments, we implement SCOD by replacing the original line 6 of Algorithm \ref{SparseCoD} with \[C_X,C_Y=\si(S_X,S_Y,\ell,1/10).\] Note that a similar strategy has also been adopted by \citet{SFD16} in the implementation of SFD.

In all experiments, each algorithm will receive two matrices $X\in \mathbb{R}^{m_x\times n}$ and $Y\in \mathbb{R}^{m_y\times n}$, and then output two matrices $B_X\in \mathbb{R}^{m_x\times \ell}$ and $B_Y\in \mathbb{R}^{m_y\times \ell}$. We adopt the approximation error $\|XY^T-B_XB_Y^T\|$ and the projection error $\|XY^T-\pi_{\bar{U}}(X)\pi_{\bar{V}}(Y)^T\|$ to measure the accuracy of each algorithm, where $\bar{U}\in\mathbb{R}^{m_x\times k},\bar{V}\in\mathbb{R}^{m_y\times k}$ and we set $k=200$. Furthermore, we report the runtime of each algorithm to verify the efficiency of our SCOD. Because of the randomness of SCOD, SFD-AMM, CS, RP and Hashing, we report the average results over $50$ runs.
\subsection{Results}
Fig. \ref{fig1} and \ref{fig2} show the results of different algorithms among different $\ell$ on the synthetic datasets. First, from the comparison of runtime, we find that our SCOD is significantly faster than COD, FD-AMM and RP among different $\ell$. Moreover, with the increase of $\ell$, the runtime of our SCOD increases more slowly than that of COD, FD-AMM and RP, which verifies the time complexity of our SCOD. Although CS and Hashing are faster than our SCOD, their accuracy is far worse than that of our SCOD. Second, in terms of approximation error and projection error, our SCOD matches or improves the performance of COD among different $\ell$. The improvement may be due to the fact that our SCOD performs fewer shrinkage than COD, which is the source of error. We note that a similar result happened in the comparison between FD and SFD by \citet{SFD16}. Third, our SCOD outperforms other baselines including FD-AMM, SFD-AMM, CS and RP.

Fig. \ref{fig3} and \ref{fig4} show the results of SCOD, COD, FD-AMM and SFD-AMM among different $\ell$ on the real datasets. The results of CS, RP and Hashing are omitted, because they are much worse than SCOD and other baselines. Compared with COD and FD-AMM, we again find that our SCOD is faster and achieves a better performance among different $\ell$. Moreover, our SCOD is slightly faster than SFD-AMM on both real datasets, and outperforms SFD-AMM in Fig. \ref{fig3}(a), \ref{fig4}(a) and \ref{fig4}(b). In Fig. \ref{fig3}(b), the performance of SCOD is very close to that of SFD-AMM.

\section{Conclusions}
In this paper, we propose SCOD to reduce the time complexity of COD for approximate multiplication of sparse matrices with the $O\left((m_x+m_y+\ell)\ell\right)$ space complexity. In expectation, the time complexity of our SCOD is $\widetilde{O}\left((\nnz(X)+\nnz(Y))\ell+n\ell^2\right)$, which is much tighter than $O\left(n(m_x+m_y+\ell)\ell\right)$ of COD for sparse matrices. Furthermore, the theoretical guarantee of our algorithm is almost the same as that of COD up to a constant factor. Experiments on both synthetic and real datasets demonstrate the advantage of our SCOD for handling sparse matrices.

\section{Acknowledgments}
This work was partially supported by the National Key R\&D Program of China (2017YFB1002201), NSFC-NRF Joint Research Project (61861146001), and the Collaborative Innovation Center of Novel Software Technology and Industrialization.

\bibliography{reference}

\clearpage
\section{Theoretical Analysis}
\subsection{Proof of Theorem \ref{thm1}}
Without loss of generality, we assume that the \emph{if} statement in Algorithm \ref{SparseCoD} is triggered $s$ times in total.

To facilitate presentations, we use $S_X^t$, $S_Y^t$, $C_X^t$, $C_Y^t$, $B_X^t$, $B_Y^t$, $D_X^t$, $D_Y^t$, $Q^t$, $Q_X^t$, $Q_Y^t$, $U^t$, $V^t$, $\Sigma^t$, $\widetilde{\Sigma}^t$ ,$\gamma_t$ to denote the values of $S_X$, $S_Y$, $C_X$, $C_Y$, $B_X$, $B_Y$, $D_X$, $D_Y$, $Q$, $Q_X$, $Q_Y$, $U$, $V$, $\Sigma$, $\widetilde{\Sigma}$, $\gamma$ after the $t$-th execution of lines 6 to 8 in Algorithm \ref{SparseCoD}, where $t=1,\cdots,s$.

Note that $B_X$ and $B_Y$ generated by Algorithm \ref{SparseCoD} are denoted by $B_X^s$ and $B_Y^s$. To bound $\left\|XY^T-B_X^sB_Y^{s,T}\right\|$, we define
\[E_1=XY^T-\sum_{t=1}^sC_X^tC_Y^{t,T}, E_2=\sum_{t=1}^sC_X^tC_Y^{t,T}-B_X^sB_Y^{s,T}\]
where $E_1+E_2=XY^T-B_X^sB_Y^{s,T}$. By the triangular inequality, we have
\begin{equation}
\label{eq1}
\begin{split}
\left\|XY^T-B_X^sB_Y^{s,T}\right\|=&\left\|E_1+E_2\right\|\leq \left\|E_1\right\|+\left\|E_2\right\|.
\end{split}
\end{equation}
Hence, we will analyze $\left\|E_1\right\|$ and $\left\|E_2\right\|$, respectively.

Combining Lemma \ref{lem2} and the union bound, with probability
$1-\sum_{j=1}^s\frac{\delta}{2j^2}\geq1-\delta$, we have
\begin{equation}
\label{eq_lemma1_eq}
\left\|S_X^tS_Y^{t,T}-C_X^tC^{t,T}_Y\right\|\leq\frac{11}{5\ell}\sum_{i=1}^{\cols(S_X^t)}\|S_{X,i}^t\|_2\|S_{Y,i}^t\|_2
\end{equation}
for all $t=1,\cdots,s$. Therefore, with probability at least $1-\delta$, we have
\begin{equation}
\label{eq2}
\begin{split}
\left\|E_1\right\|=&\left\|\sum_{t=1}^sS_X^tS_Y^{t,T}-\sum_{t=1}^sC_X^tC_Y^{t,T}\right\|\\
\leq&\sum_{t=1}^s\left\|S_X^tS_Y^{t,T}-C_X^tC_Y^{t,T}\right\|\\
\leq&\frac{11}{5\ell}\sum_{t=1}^s\sum_{i=1}^{\cols(S_X^t)}\|S_{X,i}^t\|_2\|S_{Y,i}^t\|_2\\
=&\frac{11}{5\ell}\sum_{i=1}^{n}\|X_i\|_2\|Y_i\|_2\\
\leq&\frac{11}{5\ell}\sqrt{\sum_{i=1}^{n}\|X_i\|_2^2}\sqrt{\sum_{i=1}^{n}\|Y_i\|_2^2}\\
\leq&\frac{11}{5\ell}\|X\|_F\|Y\|_F
\end{split}
\end{equation}
where the second inequality is due to (\ref{eq_lemma1_eq}) and the third inequality is due to Cauchy-Schwarz inequality.

Then, for $\left\|E_2\right\|$, we have
\begin{equation*}
\begin{split}
&\left\|E_2\right\|\\
=&\left\|\sum_{t=1}^sC_X^tC_Y^{t,T}+\sum_{t=1}^s\left(B_X^{t-1}B_Y^{t-1,T}-B_X^tB_Y^{t,T}\right)\right\|\\
=&\left\|\sum_{t=1}^s\left(D_X^{t}D_Y^{t,T}-B_X^tB_Y^{t,T}\right)\right\|\\
\leq&\sum_{t=1}^s\left\|D_X^{t}D_Y^{t,T}-B_X^tB_Y^{t,T}\right\|.
\end{split}
\end{equation*}
According to Algorithms \ref{SparseCoD} and \ref{DenseCoD}, we have \[\left\|D_X^{t}D_Y^{t,T}-B_X^tB_Y^{t,T}\right\|=\left\|\Sigma^t-\widetilde{\Sigma}^t\right\|\]
which further implies that
\begin{equation}
\label{eq3}
\begin{split}
\left\|E_2\right\|
\leq\sum_{t=1}^s\left\|\Sigma^t-\widetilde{\Sigma}^t\right\|\leq\sum_{t=1}^s\gamma_t.
\end{split}
\end{equation}
Now we need to upper bound $\sum_{t=1}^s\gamma_t$ with properties of $X$ and $Y$. First, we have
\begin{equation*}
\begin{split}
\left\|B_X^sB_Y^{s,T}\right\|_{\ast}=&\sum_{t=1}^s\left(\left\|B_X^tB_Y^{t,T}\right\|_{\ast}-\left\|B_X^{t-1}B_Y^{t-1,T}\right\|_{\ast}\right)\\
=&\sum_{t=1}^s\left(\left\|D_X^tD_Y^{t,T}\right\|_{\ast}-\left\|B_X^{t-1}B_Y^{t-1,T}\right\|_{\ast}\right)\\
&-\sum_{t=1}^s\left(\left\|D_X^tD_Y^{t,T}\right\|_{\ast}-\left\|B_X^{t}B_Y^{t,T}\right\|_{\ast}\right)
\end{split}
\end{equation*}
where $\|A\|_{\ast}$ denotes the nuclear norm of any matrix $A$.
According to Algorithms \ref{SparseCoD} and \ref{DenseCoD}, it is not hard to verify that \[\left\|D_X^tD_Y^{t,T}\right\|_{\ast}=\tr(\Sigma^t) \text{ and } \left\|B_X^{t}B_Y^{t,T}\right\|_{\ast}=\tr(\widetilde{\Sigma}^t).\]
Then, we have
\begin{equation*}
\begin{split}
&\left\|B_X^sB_Y^{s,T}\right\|_{\ast}\\
\leq&\sum_{t=1}^s\left(\left\|D_X^tD_Y^{t,T}\right\|_{\ast}-\left\|B_X^{t-1}B_Y^{t-1,T}\right\|_{\ast}\right)\\
&-\sum_{t=1}^s\left(\tr(\Sigma^t)-\tr(\widetilde{\Sigma}^t)\right)\\
\leq&\sum_{t=1}^s\left(\left\|D_X^tD_Y^{t,T}\right\|_{\ast}-\left\|B_X^{t-1}B_Y^{t-1,T}\right\|_{\ast}\right)-\sum_{t=1}^s\ell\gamma_t\\
\leq&\sum_{t=1}^s\left\|D_X^tD_Y^{t,T}-B_X^{t-1}B_Y^{t-1,T}\right\|_{\ast}-\sum_{t=1}^s\ell\gamma_t\\
=&\sum_{t=1}^s\left\|C_X^tC_Y^{t,T}\right\|_{\ast}-\sum_{t=1}^s\ell\gamma_t.
\end{split}
\end{equation*}
Furthermore, we have
\begin{equation}
\label{eq5}
\begin{split}
\sum_{t=1}^s\gamma_t&\leq\frac{1}{\ell}\left(\sum_{t=1}^s\left\|C_X^tC_Y^{t,T}\right\|_{\ast}-\left\|B_X^sB_Y^{s,T}\right\|_{\ast}\right)\\
&\leq\frac{1}{\ell}\sum_{t=1}^s\left\|Q^tQ^{t,T}S_X^tS_Y^{t,T}\right\|_{\ast}\leq\frac{1}{\ell}\sum_{i=1}^s\left\|S_X^tS_Y^{t,T}\right\|_{\ast}\\
&=\frac{1}{\ell}\sum_{t=1}^s\left\|\sum_{i=1}^{\cols(S_X^t)}S_{X,i}^tS_{Y,i}^{t,T}\right\|_{\ast}\\
&\leq\frac{1}{\ell}\sum_{t=1}^s\sum_{i=1}^{\cols(S_X^t)}\left\|S_{X,i}^tS_{Y,i}^{t,T}\right\|_{\ast}\\
&\leq\frac{1}{\ell}\sum_{t=1}^s\sum_{i=1}^{\cols(S_X^t)}\left\|S_{X,i}^t\right\|_2\left\|S_{Y,i}^{t}\right\|_{2}\\
&=\frac{1}{\ell}\sum_{i=1}^{n}\|X_i\|_2\|Y_i\|_2\\
&\leq\frac{1}{\ell}\sqrt{\sum_{i=1}^{n}\|X_i\|_2^2}\sqrt{\sum_{i=1}^{n}\|Y_i\|_2^2}\\
&\leq\frac{1}{\ell}\|X\|_F\|Y\|_F
\end{split}
\end{equation}
where the sixth inequality is due to Cauchy-Schwarz inequality. Combining with (\ref{eq1}), (\ref{eq2}), (\ref{eq3}) and (\ref{eq5}), we complete this proof.

\subsection{Proof of Lemma \ref{lem2}}
In this proof, we analyze the properties of $C_X$ and $C_Y$ that are returned by the $j$-th run of Algorithm \ref{BoostedSI}. First, we introduce the following lemma.
\begin{lem}
\label{lem1}
Let $\mathbf{x}=(x_1,x_2,\cdots,x_{m_x})\sim\mathcal{N}(0,1)^{m_x\times 1}$, $\mathbf{e}_1=(1,0,\cdots,0)\in\mathbb{R}^{m_x}$, $0<\delta<1$,
\[\pr\left[|\mathbf{e}_1^T\mathbf{x}|\leq \frac{\delta}{\sqrt{m_xe}}\|\mathbf{x}\|_2\right]\leq\delta.\]
\end{lem}
Let $U=[\mathbf{u}_1,\cdots,\mathbf{u}_{m_x}]$ denote the left singular matrix of \[C=(S_X S_Y^{T}-C_XC^{T}_Y)/\Delta\] where $UU^T=U^TU=I_{m_x}$. Because of $p=\left\lceil\log(2j^2\sqrt{m_xe}/\delta)\right\rceil$, if $\|C\|>2$, we have
\[\|(CC^T)^{p}\mathbf{x}\|_2>|\mathbf{u}_1^T\mathbf{x}|4^{p}>\|\mathbf{x}\|_2\]
as long as $|\mathbf{u}_1^T\mathbf{x}|>\frac{\delta}{2j^2\sqrt{m_xe}}\|\mathbf{x}\|_2$. Note that
\begin{align*}
&\pr\left[|\mathbf{u}_1^T\mathbf{x}|\leq\frac{\delta}{2j^2\sqrt{m_xe}}\|\mathbf{x}\|_2\right]\\
=&\pr\left[| \mathbf{u}_1^TUU^T\mathbf{x}|\leq\frac{\delta}{2j^2\sqrt{m_xe}}\|U^T\mathbf{x}\|_2\right]\\
=&\pr\left[|\mathbf{e}_1^TU^T\mathbf{x}|\leq\frac{\delta}{2j^2\sqrt{m_xe}}\|U^T\mathbf{x}\|_2\right]\\
\leq&\delta/2j^2
\end{align*}
where the inequality is due to the fact $U^T\mathbf{x}\sim\mathcal{N}(0,1)^{m_x\times 1}$ and Lemma \ref{lem1}. Therefore, when $\|C\|>2$, we have
\[\pr\left[\|(CC^T)^{p}\mathbf{x}\|_2\leq\|\mathbf{x}\|_2\right]\leq\delta/2j^2.\]
Hence, if $\|C\|>2$ but $\|(CC^T)^{p}\mathbf{x}\|_2\leq \|\mathbf{x}\|_2$, we will have
\[\left\|S_X S_Y^{T}-C_XC^{T}_Y\right\|>\frac{11\sum_{i=1}^{\cols(S_X)}\|S_{X,i}\|_2\|S_{Y,i}\|_2}{5\ell}\]
the probability of which is at most $\delta/2j^2$. We complete this proof.

\subsection{Proof of Lemma \ref{lem1}}
Let $c=\frac{\delta}{\sqrt{m_xe}}$ and $\lambda=\frac{1-m_xc^2}{2m_xc^2(1-c^2)}>0$, we have
\begin{align*}
&\pr\left[|\mathbf{e}_1^T\mathbf{x}|\leq c\|\mathbf{x}\|_2\right]\\
=&\pr\left[(c^2-1)x_1^2+c^2\sum_{i=2}^{m_x}x_i^2\geq0\right]\\
=&\pr\left[e^{\lambda(c^2-1)x_1^2+\lambda c^2\sum_{i=2}^{m_x}x_i^2}\geq1\right]\\
\leq&\mathbb{E}\left[e^{\lambda(c^2-1)x_1^2+\lambda c^2\sum_{i=2}^{m_x}x_i^2}\right]\\
=&\mathbb{E}\left[e^{\lambda(c^2-1)x_1^2}\right]\Pi_{i=2}^{m_x}\mathbb{E}\left[e^{\lambda c^2x_i^2}\right]\\
\leq&(1-2\lambda(c^2-1))^{-1/2}(1-2\lambda c^2)^{-(m_x-1)/2}\\
=&\sqrt{m_x}c\left(1+\frac{1}{m_x-1}\right)^{\frac{m_x-1}{2}}\left(1-c^2\right)^{\frac{m_x-1}{2}}\\
\leq&\sqrt{m_xe}c\\
=&\delta
\end{align*}
where the first inequality is due to Markov inequality, the second inequality is due to $\mathbb{E}\left[e^{sx_i^2}\right]=\frac{1}{\sqrt{1-2s}}$ for $i=1,\cdots,m_x$ and any $s<{1/2}$ and the third inequality is due to $(1+1/x)^x\leq e$ for any $x>0$.

\subsection{Proof of Theorem \ref{thm2}}
Note that the proof of Theorem 3 in \citet{CoD17} has already shown
\begin{align*}&\|XY^T-\pi_{\bar{U}}(X)\pi_{\bar{V}}(Y)^T\|\\
\leq&4\|XY^T-B_XB_Y^T\|+\sigma_{k+1}(XY^T).
\end{align*}
Therefore, combing with our Theorem \ref{thm1}, with probability at least $1-\delta$, we have
\begin{align*}
&\|XY^T-\pi_{\bar{U}}(X)\pi_{\bar{V}}(Y)^T\|\\
\leq&4\|XY^T-B_XB_Y^T\|+\sigma_{k+1}(XY^T)\\
\leq&\frac{64\|X\|_F\|Y\|_F}{5\ell}+\sigma_{k+1}(XY^T)\\
\leq&\sigma_{k+1}(XY^T)\left(1+\frac{64\sqrt{\sr(X)\sr(Y)}}{5\ell}\frac{\|X\|\|Y\|}{\sigma_{k+1}(XY^T)}\right)\\
\leq&\sigma_{k+1}(XY^T)(1+\epsilon).
\end{align*}
\end{document}